\documentclass{article}

 \usepackage[preprint]{nips_2018}

\usepackage[utf8]{inputenc} 
\usepackage[T1]{fontenc}    
\usepackage{hyperref}       
\usepackage{url}            
\usepackage{booktabs}       
\usepackage{amsfonts}       
\usepackage{nicefrac}       
\usepackage{microtype}      

\RequirePackage{ifpdf}
\ifpdf
  \usepackage[pdftex]{graphicx}
\else
  \usepackage[dvipdfmx]{graphicx}
\fi

\title{Reconstruction of training samples from loss functions}

\newtheorem{thm}{Theorem}[section]

\newtheorem{cor}[thm]{Corollary}
\newtheorem{prop}[thm]{Proposition}

\newtheorem{dfn}[thm]{Definition}

\newtheorem{ack}{Acknowledgments}

\author{
  Akiyoshi Sannai \\
  Center for Advanced Intelligence Project\\
  RIKEN\\
  1-4-1, Nihonbashi, Chuo, Tokyo 103-0027, Japan \\
  \texttt{akiyoshi.sannai@riken.jp} \\
}

\begin{document}

\maketitle

\begin{abstract}
  This paper presents a new mathematical framework to analyze the loss functions of deep neural networks with ReLU functions. Furthermore, as as application of this theory, we prove that the loss functions can reconstruct the inputs of the training samples up to scalar multiplication (as vectors) and can provide the number of layers and nodes of the deep neural network. Namely, if we have all input and output of a loss function (or equivalently all possible learning process),  for all input of each training sample $x_i \in \mathbb{R}^n$, we can obtain vectors $x'_i\in \mathbb{R}^n$ satisfying $x_i=c_ix'_i$ for some $c_i \neq 0$. To prove theorem, we introduce the notion of virtual polynomials, which are polynomials written as the output of a node in a deep neural network.  Using virtual polynomials, we find an algebraic structure for the loss surfaces, called semi-algebraic sets. We analyze these loss surfaces from the algebro-geometric point of view. Factorization of polynomials is one of the most standard ideas in algebra. Hence, we express the factorization of the virtual polynomials in terms of their active paths. This framework can be applied to the leakage problem in the training of deep neural networks.  The main theorem in this paper indicates that there are many risks associated with the training of deep neural networks. For example, if we have N (the dimension of weight space) + 1 nonsmooth points on the loss surface, which are sufficiently close to each other, we can obtain the input of training sample up to scalar multiplication. 
  We also point out that the structures of the loss surfaces depend on the shape of the deep neural network and not on the training samples.
\end{abstract}

\section{Introduction}
Deep learning has had great success in many fields. Deep learning model perform extremely well in computer
vision [21], image processing, video processing, face recognition
[27], speech recognition [15], and natural
language processing [1, 6, 30].
Deep learning has also been used in more complex systems that are able to play
games [14, 19, 28] or diagnose and classify diseases [2, 9, 10].\\
Along with the development of deep learning, decisions made using deep-learning principles are being implemented in a wide range of applications.
In order for deep learning to be more useful for human beings, 
it is necessary to ensure that there is no leakage of personal or confidential information in the learning process or decision-making process.
In this paper, we point out the leakage problem in deep learning. Namely, the learning process of deep learning might leak sample data. This type of phenomenon is specific to the deep-learning methods with ReLU functions.
The following example expresses the difference between linear and ReLU models.
Consider a one-dimensional least-squares model.
\[ g(a) :=  \sum_{i = 1}^m(y_i - a x_i)^2.\]
Because many $\{(x_i, y_i)| i=1, \ldots , m\}$ give the same $g(a)$ (See [25]), we cannot reconstruct $\{(x_i, y_i) |i=1, \ldots , m \}$ from $g(a)$.
However, if we consider the one-dimensional ReLU least-squares model,
\[ h(a) :=  \sum_{i =1}^m ({\rm max}(0, y_i - a x_i))^2.\]
Then, as $\{(y_i/x_i) |i=1, \ldots , m\}$ are the nonsmooth points of $h(a)$, we can obtain the nonsmooth point of $h(a)$ from $h(a)$. Hence, we can reconstruct $\{(x_i, y_i) |i=1, \ldots , m\}$ from $h(a)$ up to scalar multiplication. Namely, we obtain $\{(x'_i, y'_i )| i=1, \ldots , m \}$ satisfying $ax'_i=x_i$ and $ay'_i=y_i$ for some $a (\neq 0) \in \mathbb{R}$.\\
The main theorem in this paper shows that, if we reveal all possible learning processes, leakage of training samples can occur. As this example indicates, nonsmooth points of loss functions plays important role. We show that the set of nonsmooth points and the set induced by some algebraic structure coincides (Theorem \ref{str2}). This is a key point of this paper. Another key point is the concept of homogenous polynomials, which is used in algebraic geometry. We find the natural multidegree (layer-wise degree) in deep neural networks. We show that loss functions are essentially homogenous polynomials (virtual polynomials) of layer-wise degree. By using the theory of homogenous polynomials, we show the correspondence of the factorization of virtual polynomials and its active paths (Theorem \ref{irr}). Finally, as an application of this theorem, we show the weak reconstruction theorem of  training samples (Theorem \ref{reconst}). We also give a theoretical algorithm to obtain weak reconstruction of
training samples (Section2.5).

\subsection{Related work} 
Leakage problem: 
The leakage problem in deep learning can be a serious issue in the future and many researchers are working on this.
Since the trained model has essential information about the training sample, it is possible to extract
sensitive information from a model [3, 11, 12].
B. Hitaj, G. Ateniese, \& F. P'erez-Cruz considered this problem by using generative adversarial networks (GANs) [18]. This type of approach is suitable for images, but unsuitable for numerical data such as medical nonimage data.
For example, generating human-model data, such as height of six feet, is usually considered normal. 
However, if identity of the model is made available, then it is a leakage problem.
This is the difference between the model generating approach and the deterministic approach.

Loss surfaces: Mathematically, the learning process of deep learning is to find the local minima of loss surfaces (loss functions).
Before this paper, some researchers had analyzed the loss surfaces.
One of their aims was achieving theoretical understanding of the generalization of deep learning.
For example, K. Kawaguchi proved that any local minima of loss surfaces associated with linear neural networks was a global minimum [4,20]. J. Pennington and Y. Bahri analyzed loss surfaces using the random matrix theory [26]. In this paper, we present a new framework to analyze loss surfaces. We study the structure of loss surfaces using algebraic geometry. This approach can contribute to the theoretical understanding of the generalization of deep learning.

Algebraic geometry:
Algebraic geometry is one of the most exciting field of pure mathematics [5,8,16,24]. Furthermore, algebraic geometry frequently applied to machine learning. For example, R. Livni, et al introduced vanishing component analysis to express the algebraic (nonlinear) structure of data sets [23].  S. Watanabe applied algebraic geometry to learning theory. He proved that an invariant defined in algebraic geometry and the one defined in learning theory coincides [29]. He also related these invariants value to zeta functions. When we treat polynomials, algebraic geometry is a powerful tool to consider them.

\subsection{Contribution}

We discuss the loss functions of fully connected deep neural networks with square losses.
Basically, all notations are taken from the deep learning book by Goodfellow, et al [13].
Let $L$ be the number of layers. We do not use the notion of "hidden layer" for the consistency of the other definitions.
We denote the weight parameters by
$w \in \mathbb{R}^N$, which consists of the entries of the parameter matrices corresponding to each layer : $W_{L-1} \in
\mathbb{R}^{d_{L}\times d_{L-1}}, \ldots, W_k \in \mathbb{R}^{{d_{k+1}}\times d_{k} }, \ldots , W_1 \in \mathbb{R}^{{d_2} \times d_1}$. Here, $d_k$ represents the width of the $k$-th layer,
where the first layer is the input layer and the $L$-th layer is the output layer. 
We use $(i, k)$-node to indicate the $i$-th node in the $k$-th layer. We denote its output as $x_i^{(k)}$ and pre-output as $z_i^{(k)}$, namely $x_i^{(k)}={\rm max}(0, z_i^{(k)})$ and
\[
W_k \left[
    \begin{array}{c}
      x_1^{(k)} \\
      x_2^{(k)} \\
      \vdots \\
      x_{d_k}^{(k)}
    \end{array}
  \right]
= \left[
    \begin{array}{c}
      z_1^{(k+1)} \\
      z_2^{(k+1)}\\
      \vdots \\
      z_{d_{k+1}}^{(k+1)}
    \end{array}
  \right]
.
\]

 We simply denote $x_i^{(1)}$ by $x_i$. We denote the output by $F_w: \mathbb{R}^{d_1} \to \mathbb{R}^{d_L}$,
namely,

\[
 F_w \left( \left[
    \begin{array}{c}
      x_1  \\
      x_2 \\
      \vdots \\
      x_{d_1}
    \end{array}
  \right] \right)
= \left[
    \begin{array}{c}
     z_1^{(L)}  \\
     z_2^{(L)} \\
      \vdots \\
      z_{d_L}^{(L)} 
    \end{array}
  \right].
\]

Let $\Omega=\{(a_i, b_i) |\ i=1, \ldots ,M\} \subset \mathbb{R}^{d_1} \times \mathbb{R}^{d_L}$ be a training sample set.
Then, we define the loss function as follows,
\[ E(w)= \sum_{(a_i, b_i)\in \Omega} \frac{1}{2}  \|b_i-F_w(a_i)\|^2 ,\]
where$ \| \cdot \|$is the Frobenius norm.
The main theorem of this paper is given below.
\begin{thm}
Let $E(w)$ be the loss function of deep neural network with ReLU functions. Assume we can obtain all input and output of $E(w)$. Then we can obtain $\{a'_i\}$ satisfying $c_i a'_i=a_i$ for some $c_i (\neq 0) \in \mathbb{R}$, number of layers, and number of nodes in each layer. 
\end{thm}

This theorem means that the input and output of the loss function $E(w)$ can reconstruct the input of the training samples up to scalar multiplication. In other word, if we can obtain all possible training process of deep learning, $\{a_i\}$ is reconstructed as $\{a'_i\}$. In general, $a'_i$ is not equal to $a_i$. However,  if we obtain a entry of $a_i$, we can specify $c_i$ in the Theorem 1.1. Hence, we can obtain $a_i$. This indicates that it carries many risks to reveal the training process of deep learning. Hence, we need to conceal the value of loss functions to protect training samples.
We can provide a stronger statement after proper mathematical preparation (See Theorem \ref{reconst}).

Note that Theorem 1.1 can be generalized as follows. First, we can add
any smooth function r(w) to the loss function E(w) as a regularization
term. Second, we can change the activation function to any piecewise
linear function such as Leaky ReLU, Maxout, and LWTA [17,22]. For simplicity,
in this paper, we only treat the simplest case.

\section{Mathematical results}
In this section, we prepare definitions and theorems to prove the main result.
Our focus is on the loss surfaces that are defined by
\[ X = \{ (w,y) \in \mathbb{R}^N \times \mathbb{R} \  | \ y= E(w)\} \]
,where $E(w)$ is the loss function defined above.
From the view point of deep learning, we are interested in the local minima of loss functions.
We provide some mathematical frameworks from algebraic geometry. This is a new method to analyze the loss surfaces, which can contribute to the theoretical understanding of generalization. For the standard notations in algebraic geometry, we refer to [8,16].
First, we define semi-algebraic sets from a field of pure mathematics, algebraic geometry. 
Let $X$ be a subset of $\mathbb{R}^N$. $X$ is said to be a semi-algebraic set if $X$ is defined by the polynomials $f_i = 0$ and
$g_j>0$ and the finite union of them. If $X$ is a semi-algebraic set, we can state that $f_i$ is a defining equation of $X$ and $g_i$ is a defining inequation of $X$.
For other notations in semi-algebraic geometry, we refer to [7]. The following theorem points out that the loss surfaces are semi-algebraic sets.

\begin{thm}[Structure theorem 1]\label{str1}
Let $X$ be a loss surface of a square loss function. Then, $X$ is a semi-algebraic set of codimension 1.
\end{thm}

\begin{figure}
\center
\includegraphics[width=10cm]{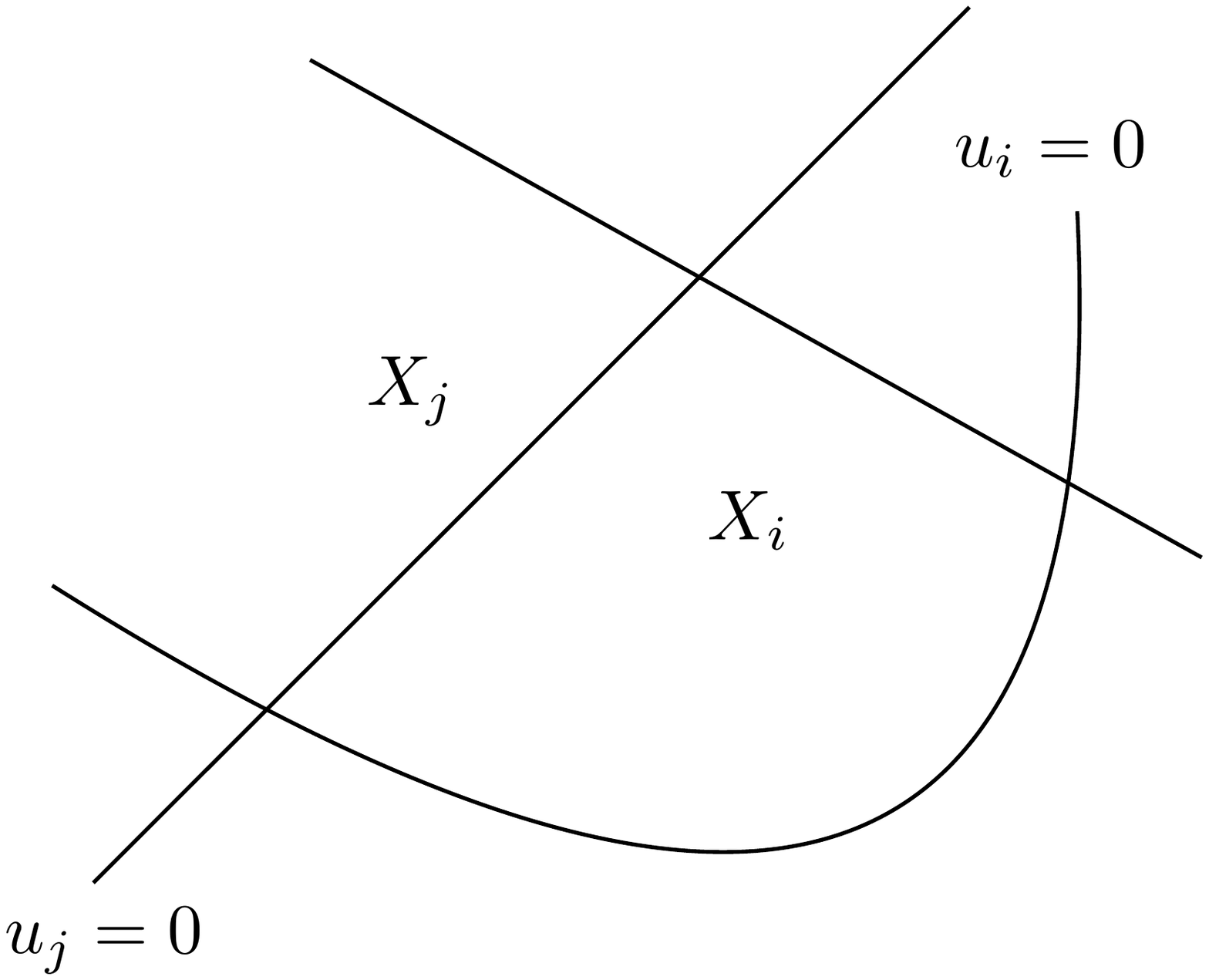}
\caption{Decomposition of a loss surface}
\label{fig:test}
\end{figure}

Figure 1 indicates the meaning of the theorem. The polynomial $u_i$ divides the loss surface into subsurfaces $X_k$ and each $X_k$ is defined by a (fixed) polynomial. We can see the precise description of $u_i$ later.This theorem allows us to use algebraic geometry for analyzing loss surfaces. 
The second theorem is about the decomposition of $X$ as a semi-algebraic set.
To describe it, we define virtual polynomials, which are functions written as the outputs of nodes.

\subsection{Virtual polynomials}
The concept of virtual polynomials plays an important role in this paper.

\begin{dfn}
Fix an input $x$ and weight $w$ on the fixed deep neural network.
An $(i,k)$-node is said to be active if its output $x_i^{(k)}$ is positive. We define the set \[ \{[(i, k, q) \mid q= {\rm positive\ or\ negative}\} \] to be the ReLU activation set.
\end{dfn}

When we mention just ReLU activation set, it is just a formal pair of a node and its activations. Hence, it is irrelevant whether it is realized by an input and a weight.
When we have a ReLU activation set, it induces a deep linear network. We define virtual polynomials by using them.

\begin{dfn}
Fix an input $x$. A weight valuable polynomial $u$ is defined to be a virtual polynomial of type $(i, k)$ if $u= x_i^{(k)}$, where $x_i^{(k)}$ is the output of the $i$-th node in the $k$-th layer in the deep linear network induced by some ReLU activation set. We simply define $u$ as a virtual polynomial if $u$ is a virtual polynomial of type $(i,k)$ for some ReLU activation set and some $(i,k)$.
\end{dfn}
See Figure 2. The virtual polynomials of type $(1, 3)$ in this neural network are 
\[
\{\omega_1\omega_5x_1 +
\omega_2\omega_6x_1 +\omega_3\omega_5x_2 +\omega_4\omega_6x_2, 
\omega_1\omega_5x_1 +\omega_3\omega_5x_2,
\omega_2\omega_6x_1 +\omega_4\omega_6x_2,
0\}.
\]

The corresponding ReLU activation sets are
\[ \{(1,1,{\rm active}),(1,2, {\rm active}),(2,1,{\rm active}), (2,2,{\rm active})\} \]
\[ \{(1,1,{\rm active}),(1,2, {\rm active}),(2,1,{\rm active}), (2,2,{\rm negative})\} \]
\[ \{(1,1,{\rm active}),(1,2, {\rm active}),(2,1,{\rm negative}), (2,2,{\rm active})\} \]
\[ \{(1,1,{\rm active}),(1,2, {\rm active}),(2,1,{\rm negative}), (2,2,{\rm negative})\}.\]
\begin{figure}
\center
\includegraphics[width=10cm]{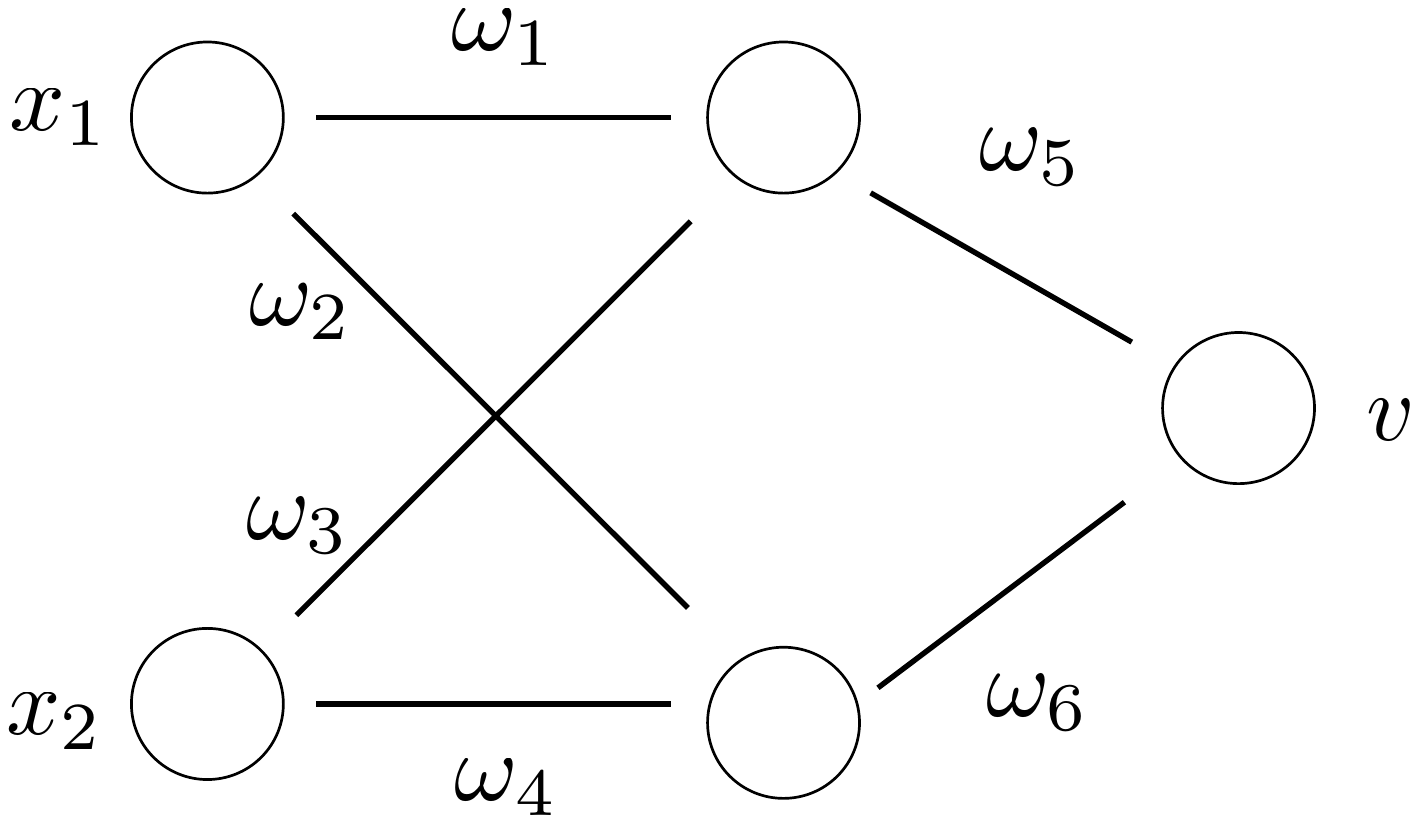}
\caption{Neural network}
\label{fig:test}
\end{figure}
If we fix a ReLU activation set, we have a virtual polynomial.
However, even if we fix the virtual polynomial, the ReLU activation set that provides the virtual polynomial is not unique.
For example, $\{(1,1,{\rm negative}),(1,2, {\rm negative}),(2,1,{\rm negative}), (2,2,{\rm negative})\}$ give $0$ as a virtual polynomial in the example above.\\
Now, we can state the second theorem.

\begin{thm}[Structure theorem 2]\label{str2}
Let $X$ be a loss surface of a square loss function. Let ${\rm Sing}(X)$ be the set of nonsmooth points on $X$. Then,
\begin{itemize}
\item The shortest decomposition (the decomposition that we cannot reduce by defining inequations) is given by ${\rm Sing}(X)$.
\item ${\rm Sing}(X)$ is purely codimension 1 in $X$ (See [3]) and is locally defined by a virtual polynomial.
\item ${\rm Sing}(X)$ is a semi-algebraic set.
\end{itemize}
\end{thm}

This indicates that ${\rm Sing}(X)$ is a natural set from not only the differential-geometric view but also the algebro-geometric view. 
By this theorem, ${\rm Sing}(X)$ is locally defined by some virtual polynomials. Hence, from the algebro-geometric view point, we need to know the irreducible decomposition of virtual polynomials to obtain the geometric structure of ${\rm Sing}(X)$. 

\subsection{Irreducibility of polynomials}
In this section, we review factrization of polynomials. We first define the irreducibility of polynomials.
Let $f$ be a polynomial with real coefficients and $n$ valuables.
$f$ is said to be irreducible if we cannot write $f$ as a product of two non-constant polynomials.
Namely,
\[
f=gh \Rightarrow  g {\rm \ or \ } h {\rm \ is \ constant} .
\]
It is well-known that polynomials have an irreducible decomposition[5,8,10].
Namely, let $f$ be a polynomial with real coefficients and $n$ valuables.
Then, $f$ has a unique decomposition of the following form.

\[
f=f_1\cdots f_n
\]
,where $f_i$ is an irreducible polynomial with real coefficients and each $f_i$ is unique up to constant multiplication. 
We define $f_i$ above as an irreducible component of $f$. 
In Section 2.4, we give the irreducible decomposition of virtual polynomials (See Theorem \ref{irr}).

\subsection{Homogenous polynomials}

In this subsection, we review the concepts of homogenous polynomials and multidegree.
Let $x_1, \ldots , x_n$ be valuables.
Multidegree of each $x_i$ is defined as an element in $\mathbb{Z}^n$.
For any monomial $m=x_1^{a_1} \cdots , x_n^{a_n}$, we define
${\rm deg}\ (m) = \sum_i a_i{\rm deg}\ (x_i)$

A deep neural network induces natural multidegree.
\begin{dfn}
Let ${w_{(i,j)}^{(k)}}$ be the weight valuable on the path passing from the $i$-th node in the $k$-th layer to the $j$-th node in the $k+1$-th layer. Then, we define
\[
{\rm deg}({w_{(i,j)}^{(k)}}) = (0, 0, \ldots, 0, 1, 0, 0, \ldots, 0),
\]
where $1$ exists in the $k$-th entry. We call this multidegree layer-wise degree.
\end{dfn}
Fix multidegree. A polynomial $f$ is said to be homogenous if any monomial appearing in $f$ has the same multidegree. In this case, we define
${\rm deg}(f)= {\rm deg}(m),$ 
where $m$ is a monomial appearing in $f$. ${\rm Deg}(m)$ does not depend on the choice of $m$. It is well-known that any irreducible component of homogenous polynomial is homogenous (See [5,8,24]). 

We can see an example of layer-wise degree in Figure 2. The layer-wise degree of this neural network is 
\[
{\rm deg}(w_i)=(1,0)    \  \ \  (i=1,2,3,4) \ \ \
\ \ {\rm deg}(w_i)=(0,1)   \  \ \  (i=5, 6)\\
\]

The following theorem points out the features of virtual polynomials.
\begin{thm}\label{deg}
Virtual polynomials of type $(i, k)$ are homogenous polynomials of layer-wise degree with ${\rm deg}(f)=(1,  \ldots ,1, 0, \ldots ,0)$, where 1 exists from the first entry to the $k$-th entry .
\end{thm}

\subsection{Irreducible decomposition theorem}
In this subsection, we give the necessary and sufficient conditions for the irreducible decomposition of virtual polynomials.

\begin{dfn}
Let $P$ be a ReLU activation set of fixed input $x_0$ and weight $w_0$. Then, a $P$-active neural network is a subneural network, which consists of $P$-active nodes and the paths between them.
\end{dfn}

An example of $P$-active neural networks is given below.
See Figure 2 and 3. We can regard the neural net in Figure 2 as a sub neural network of the one in Figure 2. Assume that $v$ in Figure 2 is negative for some input and weight and the earlier output was positive. Then, with this ReLU activation set $P$, the $P$-active neural network is equal to the one in Figure 2.

\begin{thm}[Irreducible decomposition theorem]\label{irr}
Let $P$ be a ReLU activation set of fixed input $x_0$ and weight $w_0$.
Let $u$ be a virtual polynomial of type $(i, L)$ induced by $P$.
Then, $u=g_1\cdots g_n$ if and only if the $P$-active neural network has $n-1$ layers such that there is a unique node in the layer. Furthermore, we can write $g_i$ as the output of the subneural network which starts from a unique node and ends at the next unique node.
\end{thm}

A typical example of Theorem \ref{irr} is given below.

 Let $u$ be a virtual polynomial with a $P$-active neural network (Figure 3). Then, we have the following irreducible decomposition of $u$
 \[
 u=(\omega_1\omega_5x_1 +\omega_2\omega_6x_1 +\omega_3\omega_5x_2 +\omega_4\omega_6x_2)
 (\omega_7\omega_9+\omega_8\omega_{10}).
\]
The first component of the decomposition corresponds to the output of the node in the third layer. The second component of the decomposition corresponds to a function that starts from the third layer and ends at the output. Hence, the theorem tells us the irreducible decomposition of the virtual polynomials from its active node.
\begin{figure}
\center
\includegraphics[width=10cm]{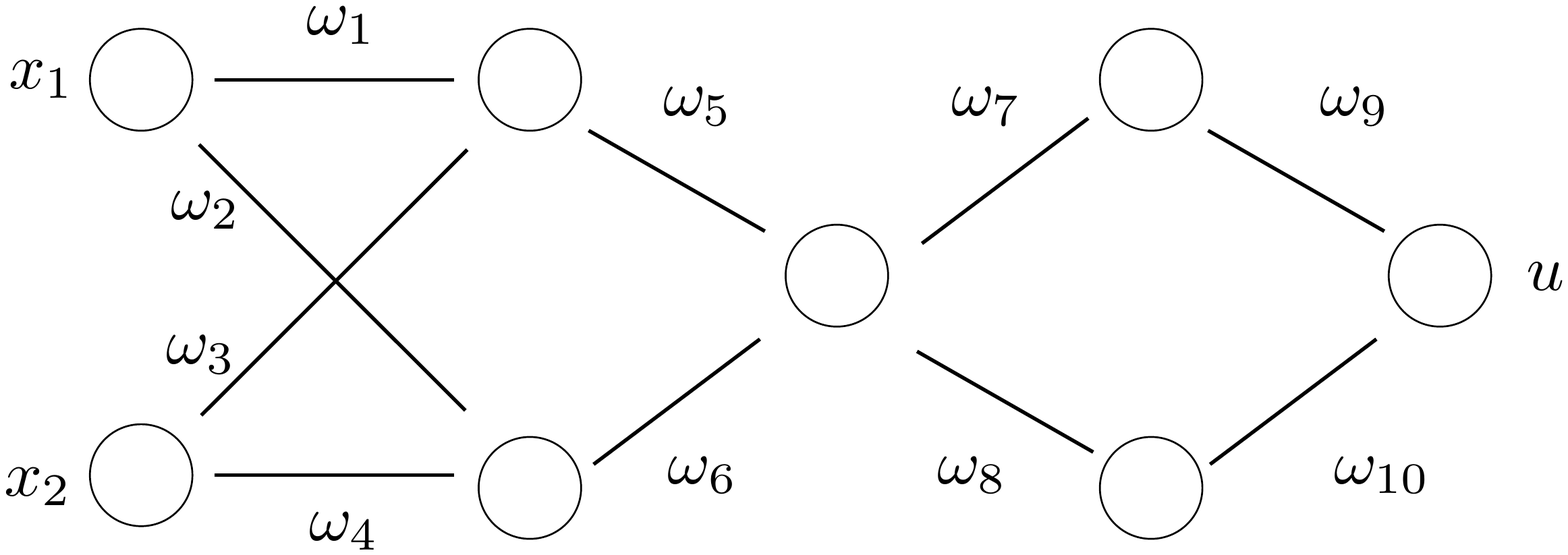}
\caption{P-active neural network}
\label{fig:test}
\end{figure}

We can see that $u$ in this example is realized by an input $x$ and weight $w$.
Hence, we can see that $u$ is one of the defining equations of ${\rm Sing}(X)$. The decomposition implies that $u=0$ if and only if 
\[
\omega_1\omega_5x_1 +\omega_2\omega_6x_1 +\omega_3\omega_5x_2 +\omega_4\omega_6x_2=0
\]
or
 \[\omega_7\omega_9+\omega_8\omega_{10}=0.\]
This means that $\omega_7\omega_9+\omega_8\omega_{10}$ is the defining equation of ${\rm Sing}(X)$. However,  $\omega_7\omega_9+\omega_8\omega_{10}$ does not depend on the training samples.
Hence, the loss surfaces have differential geometric structures, which are independent of the training samples. Suitable algorithms using such structures can be developed.

\begin{prop}
${\rm Sing}(X)$ has irreducible components, which do not depend on the training samples.
\end{prop}

\begin{cor}
Linear components of virtual polynomials are weight parameters or come from the second layer.
\end{cor}

We state the main result of this paper. Note that, if we know the input and output of the loss functions, we know the defining equations and the defining inequations of ${\rm Sing}(X)$.
\begin{thm}[Weak reconstruction theorem]\label{reconst}
${\rm Sing}(X)$ reconstructs the number of layers, number of nodes in the layers, and training samples (not equal to unit vector) up to scalar multiplication. Namely, a vector $a'_i\in \mathbb{R}^n$ satisfying $a_i=c_ia'_i$ for some $c_i \neq 0$ is reconstructed for all input of each training sample $a_i \in \mathbb{R}^n$.
\end{thm}
 In this theorem, we need infinite points on the loss surface. However, if we assume that these points are sufficiently close to each other, we can reconstruct the input of a training sample.
\begin{prop}\label{pr}
Assume that we have $N+1$ nonsmooth points on the loss surface,
which are not smooth and are sufficiently close to each other, then, we can obtain a vector $a'\in \mathbb{R}^n$ satisfying $a_i=ca'$ for some $i$ and $c \neq 0$.
\end{prop}

\subsection{An algorithm to reconstruct training samples}
We give a theoretical algorithm to obtain weak reconstruction of
training samples. The algorithm requires long time, but terminates in
a finite time.

We estimate the degree of defining equations from the dimension of the
weight space.
Then, we take random weight $w_1$ and obtain the defining equation $y
= f(w)$ around $w_1$ by taking finitely many points near $w_1$.
After that, we find an adjacent division and its defining equation $y
= g(w)$ by taking random points $w_2$ around $w_1$ and comparing the
values of $f(w_2)$ and $E(w_2)$. Here, the intersection of these two
equations is an irreducible component of Sing(X). In other word, $f(w)
- g(w)$ is a virtual polynomial.
We repeat this procedure until we find all the divisions.

\section{Sketch of proofs}\label{Preliminaries}


We give sketches of the proofs in this section. We complete the proofs in the appendix.

\subsection{Proof of Theorem \ref{str1}}
Let $u_{(i, k)}$ be a virtual polynomial at (i,k).
Then, put
\[W_{u} =\{w \in \mathbb{R}^N \mid u_{(i, k)}(w)=0\} .\]
Since $W_{u}$ is defined by a  single polynomial, $W_{u} $ divides the weight space into two areas defined by inequalities.
Add all $W_{u}$ to the weight space. Then, the space is divided into many areas defined by inequalities.
Fix an input $x$ and take two weights $w_1$ and $w_2$. If $w_1$ and $w_2$ belong to the same area, the ReLU activation set associated with $(x, w_1)$ and the one associated with $(x, w_2)$ are the same. This implies that, if the weights are in the same area, any entry of $F_w$ is a polynomial. Hence the loss function is a polynomial. This means that the loss surface is a semi-algebraic set.

\subsection{Proof of Theorem \ref{reconst} and Proposition \ref{pr}}
By the assumption, we may assume that we know the defining equations of ${\rm Sing}(X)$, we can pick up the linear polynomials in it. We show that, if the linear polynomial is not a weight parameter,
the coefficients are equal to the input of some samples up to scalar multiplication.
First, we remark that the coefficients of the virtual polynomials in the second layer are equal to the inputs of some samples up to scalar multiplication.
Since we can see that the defining polynomials of ${\rm Sing}(X)$ include virtual polynomials in the second layer,
it is proof enough that any linear polynomials appearing in the defining polynomials are virtual polynomials in the second layer. If we assume that a linear polynomial appears in the defining polynomials of ${\rm Sing}(X)$, it will be the irreducible component of a virtual polynomial. By Theorem \ref{irr}, we can see that it is the virtual polynomial of the second layer or weight parameter. This is because, if a linear polynomial that is not a virtual polynomial of the second layer appears as an irreducible component of the virtual polynomial, it must start from the $l$-th layer with one active node and end at the $l+1$-th layer with one active node. This is a weight parameter.
Hence, we reconstruct the input of a sample up to scalar multiplication and the weights on the paths from the first layer to the second layer.  We can find a quadric polynomial in the defining equation of ${\rm Sing}(X)$, which contains the weights on the paths from the first layer to the second layer. Then, the remaining weights are the weights on the paths from the second layer to the third layer. Inductively, we can reconstruct the number of nodes and layers.

\section{Conclusion}
In this paper, we presented a new mathematical framework based on algebraic geometry and some new concepts including virtual polynomials. 
Using these, we proposed a structure theorem for loss surfaces, an irreducible decomposition theorem for virtual polynomials. The main contribution of this paper was the reconstruction theorem for samples. Namely, the training process of deep learning could leak information of samples. While this fact is important on its own,
the proposed framework contributes more. This framework enables researchers in the fields of machine learning and algebraic geometry to pursue research on deep learning. 
We will be able to discover new algorithms on security issues, from this framework.
In addition, we may be able to find an efficient training algorithm on deep learning from this framework.
 More theoretical understanding of deep learning is required, but there is also a possibility of contributing to this.

\begin{ack}
The author would like to thank Prof. Masashi Sugiyama and Kenichi Bannai for giving the opportunity to study machine learning at RIKEN AIP.  The author would like to thank Prof. Jun Sakuma and Takanori Maehara for carefully reading the draft and offering valuable advice. The author would like to thank Prof. Shuji Yamamoto and Sumio Watanabe for their fruitful discussion. The author was partially supported by JSPS Grant-in-Aid for Young Scientists (B) 16K17581.
\end{ack}

\small

[1] A. Abdulkader, A. Lakshmiratan, and J. Zhang. (2016) Introducing
DeepText: Facebook's text understanding engine. https://tinyurl.com/
jj359dv

[2] A. Cruz-Roa, J. Ovalle, A. Madabhushi, and
F. Osorio. (2013)
 A deep learning architecture for image
representation, visual interpretability and automated basal-cell carcinoma cancer
detection. {\it In International Conference on Medical Image Computing and ComputerAssisted
Intervention. Springer Berlin Heidelberg, 403--410.}

[3] G. Ateniese, L. V Mancini, A. Spognardi, A. Villani,
D. Vitali, \& G. Felici. (2015)
Hacking smart machines with
smarter ones: How to extract meaningful data from machine learning classifiers.
{\it International Journal of Security and Networks 10, 3, 137--150.}

[4] P. Baldi \ \& K. Hornik. (1989)
Neural networks and principal component analysis: Learning
from examples without local minima. 
{\it Neural networks, 2(1), 53--58 .}

[5] W. Bruns \ \& H. J. Herzog. (1998)
{\it Cohen-Macauley rings}
Cambridge University Press

[6] R. Collobert, J. Weston, L. Bottou, M. Karlen, K. Kavukcuoglu,
and P. Kuksa. (2011)
 Natural language processing (almost) from scratch.
{\it Journal of Machine Learning Research 12, Aug (2011), 2493--2537.}

[7] M. Coste. (2002)
An introduction to semialgebraic geometry. Tech. rep., 
{\it Institut de Recherche
Mathematiques de Rennes }

[8] D. Cox, J. Little, and D. O’Shea. (1992)
{\it Ideals, Varieties, and Algorithms: An
Introduction to Computational Algebraic Geometry and Commutative
Algebra.}
Springer.

[9] DeepMind. 2016. DeepMind Health, Clinician-led. Patient-centred. (2016). https:
//deepmind.com/applied/deepmind-health/

[10] R. Fakoor, F. Ladhak, A. Nazi, and M. Huber. (2013) Using deep
learning to enhance cancer diagnosis and classification. {\it In The 30th International
Conference on Machine Learning (ICML 2013),WHEALTH workshop}

[11] M. Fredrikson, S. Jha, and T. Ristenpart. (2015)
Model inversion
attacks that exploit confidence information and basic countermeasures. 
{\it In Proceedings
of the 22nd ACM SIGSAC Conference on Computer and Communications
Security. ACM, 1322--1333.}

[12] M. Fredrikson, E. Lantz, S. Jha, S. Lin, D. Page, and T. Ristenpart. (2014) 
Privacy in pharmacogenetics: An end-to-end case study of
personalized warfarin dosing. 
{\it In 23rd USENIX Security Symposium (USENIX
Security 14). 17--32.}

[13] I. Goodfellow, Y. Bengio \ \& A.Courville. (2016)
{\it Deep Learning.}
MIT Press.

[14] Google DeepMind. 2016. AlphaGo, the first computer program to ever beat a
professional player at the game of GO. (2016). https://deepmind.com/alpha-go

[15] A. Graves, A. Mohamed, and G. Hinton. (2013)
Speech
recognition with deep recurrent neural networks. 
{\it In 2013 IEEE international
conference on acoustics, speech and signal processing. IEEE, 6645--6649.}

[16] R. Hartshorne. (1977)
{\it Algebraic geometry, }
Springer-Verlag, New York, Graduate Texts in Mathematics,
No. 52

[17] K. He, X. Zhang, S. Ren \ \& J. Sun.  (2015)
Delving Deep into Rectifiers: Surpassing Human-Level Performance on ImageNet Classification
{\it IEEE International Conference on Computer Vision }

[18] B. Hitaj, G. Ateniese, \ \& F. Perez-Cruz. (2017)
{\it Deep models under the GAN: information leakage from collaborative
deep learning. CoRR, abs/1702.07464.}

[19] M. Lai. (2015) Giraffe: Using deep reinforcement learning to play chess.
{\it arXiv preprint arXiv:1509.01549 (2015).}

[20] K. Kawaguchi. (2016)
Deep learning without poor local minima,
{\it In Advances In Neural Information Processing Systems, pp. 586--594, 2016}.

[21] Y. LeCun, K. Kavukcuoglu, C. Farabet, et al. (2010)
Convolutional
networks and applications in vision.{\it In ISCAS. 253--256.}

[22] Z. Liao\ \& G. Carneiro.
On the Importance of Normalisation Layers in Deep Learning with Piecewise
Linear Activation Units,
{\it arxiv1508.0033}

[23] R. Livni, D. Lehavi , S. Schein, H. Nachlieli, S. Shalev-Shwartz\ \& A. Globerson.  (2013)
Vanishing Component Analysis
{30th International Conference on Machine Learning. }

[24] H. Matsumura.
{\it Commutative Ring Theory}
Cambridge Studies in Advanced Mathematics

[25] M. Marshall. (2008)
 Positive Polynomials and Sums of Squares ,
{\it Mathematical Surveys and Monographs
Volume: 146}

[26] J. Pennington \ \& Y. Bahri. (2017)
 Geometry of Neural Network Loss Surfaces via Random Matrix Theory
 {\it Proceedings of the 34th International Conference on Machine Learning, PMLR 70:2798--2806.}
 
[27] Y. Taigman, Ming Yang, Marc’Aurelio Ranzato, and Lior Wolf. (2014)
  DeepFace:
Closing the Gap to Human-Level Performance in Face Verification. 
{\it In Proceedings
of the 2014 IEEE Conference on Computer Vision and Pattern Recognition, 1701--1708.}
 
 [28] V. Mnih, K. Kavukcuoglu, D. Silver, A. Graves, I.
Antonoglou, D. Wierstra, and M. Riedmiller. (2013) Playing atari with deep
reinforcement learning. {\it arXiv:1312.5602 (2013).}
 
 [29] S. Watanabe. (2009)
{\it Algebraic Geometry and Statistical Learning Theory}
Cambridge University Press
 
[30] X. Zhang and Y. LeCun. (2016) Text Understanding from Scratch.
{\it arXiv preprint arXiv:1502.01710v5 (2016).}

{\large Appendix of Reconstruction of training samples from loss functions.}

\subsection{Proof of Theorem 2.1}
Let  $a_p$ be an input of a training sample.
Let $V_p$ be the set of virtual polynomial of the input $a_p$.
We show that $V=\bigcup_{p=1}^M \{ u_{(i, k)}^{(p)}(w)>0,u_{(i, k)}^{(p)}(w)<0 | \ u_{(i, k)}^{(p)} \in V_p \} $ and equations define the loss surface.
Put
\[W(u_{(i, k)}^{(p)}) =\{w \in \mathbb{R}^N \mid u_{(i, k)}^{(p)}(w)=0\} .\]
\[W^+(u_{(i, k)}^{(p)})=\{w \in \mathbb{R}^N \mid u_{(i, k)}^{(p)}>0\} .\]
\[W^-(u_{(i, k)}^{(p)}) =\{w \in \mathbb{R}^N \mid u_{(i, k)}^{(p)}<0\} .\]
Take two weights $w_1$ and $w_2$ from $(\bigcap_{u_{(i, k)}^p \in V^+}{W^+(u_{(i, k)}^p)}) \bigcap (\bigcap_{u_{(i, k)}^p \in V^-}{W^-(u_{(i, k)}^p)})$, where $V^+ \cup V^- =V$. The ReLU activation set associated with $(a_k, w_1)$ and the one associated with $(a_p, w_2)$ are the same. This implies that, if the weights $w$ are in the same area, the entry of $F_w(a_k)$ is a polynomial. Hence, the loss function is a fixed polynomial $f(w)$ in this area. The loss surface is defined by $y=f(w)$ in $(\bigcap_{u_{(i, k)}^p \in V^+}{W^+(u_{(i, k)}^p)}) \bigcap (\bigcap_{u_{(i, k)}^p \in V^-}{W^-(u_{(i, k)}^p)})$. If the weights are in $(\bigcap_{u_{(i, k)}^p \in V'}{W(u_{(i, k)}^p)})$, put $V^*=V-V'$. In this case, we can use the same discussion for $V^*$. This means that the loss surface is a semi-algebraic set.

\subsection{Proof of Theorem 2.4}
The proof of Theorem 2.1 implies that $W(u_{(i, k)}^{(p)}) $ gives a decomposition. Put $\tilde{W}=(\bigcap_{u_{(i, k)}^p \in V^+}{W^+(u_{(i, k)}^p)}) \bigcap (\bigcap_{u_{(i, k)}^p \in V^-}{W^-(u_{(i, k)}^p)})$, where $V^+ \cup V^- =V$.
We see that, the loss surface is smooth in the domain $\tilde{W}$.
Again, by the proof of Theorem 2.1, the loss function is a polynomial in $\tilde{W}$.
Then, the loss surface is defined in the form of $y = f(w)$, where $f(w)$ is a polynomial.
By the Jacobian criterion, we see that the loss surface is smooth in $\tilde{W}$ for any $V^+, V^-$.\\
We then claim that we can erase the virtual polynomial ${u_{(i, k)}^p}$ from the defining inequalities if and only if $W(u_{(i, k)}^{(p)})$ consist of smooth points. 
Take a point $x$ in $\tilde{W}$. If the point is smooth, we can take the Taylor expansion of $E(w)$. Since $E(w)$ is a polynomial at a point in the neighborhood of $x$, the Taylor expansion of $E(w)$ will be a polynomial.
Hence, we can erase ${u_{(i, k)}^p}$ from the defining inequations.
Conversely, we assume that we can erase ${u_{(i, k)}^p}$ from the defining inequations. Since the loss surface around any point in $W({{u_{(i, k)}^p}})$ is defined by a polynomial, the point is smooth.
Finally, we show that Sing$(X)$ is also a semi-algebraic set of codimension 1 in $X$.
Note that Sing$(X)$ is locally of the form $W(u_{(i, k)}^{(p)}) $. This implies that Sing$(X)$ is codimension 1 in $X$. We consider the decomposition discussed in the proof of Theorem2.1 again. In each domain, it is fixed that $W({{{u'}_{(i, k)}^p}})$ is singular or not, because the function ${{{u'}_{(i, k)}^p}}$ is a polynomial in each domain. This implies that Sing$(X)$ is a semi-algebraic set.

\subsection{Proof of Theorem 2.6}
By the construction of the virtual polynomial, the weights of each layer appear precisely once in each monomial. This implies that the layer-wise degree is equal to $(1, \ldots , 1, 0, \ldots, 0)$.

\subsection{Proof of Theorem 2.8}
We prove the theorem for any connected deep neural networks by induction on the number of layers.
If $n=1$, the statement is clear.
Assume $n>1$.
First, by Theorem2.15, virtual polynomials of type $(i,L)$ are a homogenous polynomials of layer-wise degree $(1, 1, \ldots ,1)$ and the layer-wise degree is realized by assigning the degree $(0, 0, \ldots, 0, 1, 0, 0, \ldots, 0)$ to the weights on the paths passing from the $l$-th layer to the $l+1$-th layer, where $1$ exists in the $l$-th entry. 
Assume that $u=g_1\cdots g_n$. Then, by a general theory of commutative algebra, $g_i$ are homogenous polynomials of layer-wise degree. We have
\[\sum_{i=1}^n {\rm deg}(g_i)={\rm deg}(u)= (1, 1, \ldots ,1) .\]
We may assume that from the first entry to the $l$-th entry, the entry of ${\rm deg}(g_1)$ is 1 and $l+1$-th entry of ${\rm deg}(g_1)$ is zero . Hence, we may also assume that the $l+1$-th entry of ${\rm deg}(g_2)$ is 1.
Let ${w_{(i,j)}^{(l)}}$ be the weight on the path passing from the $i$-th node in the $l$-th layer to the $j$-th node in the $l+1$-th layer. There are the monomials containing ${w_{(i,j)}^{(l)}}$ in $g_1$ and monomials containing ${w_{(r,s)}^{(l+1)}}$ in $g_2$ by the layer-wise degree. After the construction of virtual polynomials, there will be no monomials in $u$ containing  ${w_{(i,j)}^{(l)}}{w_{(r,s)}^{(l+1)}}$ if $j\neq r$. However, $u=g_1\cdots g_n$ implies that $u$ has monomials containing  ${w_{(i,j)}^{(l)}}{w_{(r, s)}^{(l+1)}}$ for all $i,j,r,s$.
This implies that the $l+1$-th layer of $P$-active neural network has a unique node. 
Hence, after the construction of virtual polynomials, we obtain $u=g'_1u'$ , where $g'_1$ is the output of the unique node in $l+1$-th layer. By the general theory of commutative algebra (uniqueness of the decomposition),  we obtain $g_1=g'_1$.
Since we can regard $u'$ is a virtual polynomial of the subneural network which starts from $l+1$-th layer induced by the output of $l+1$-th layer and the same weights, the theorem holds for $u'$ by inductive hypothesis. This completes the proof of the first statement in Theorem 2.18. The remaining claim follows from the construciton of $g'_1$.

\subsection{Proof of Corollary 2.10 and Theorem 2.11}
By the assumption, we may assume that we know the defining equations of ${\rm Sing}(X)$, we can pick up the linear polynomials in it. Let $L_i= \sum_{k=1}^m a_k'w_k$ is linear polynomials in it.We show that, if $L_i$ is not a weight parameter,
$(a_k')$ is equal to the input of some samples up to scalar multiplication.
First, we remark that the coefficients of the virtual polynomials in the second layer are equal to the inputs of some samples. Since we can see that the defining polynomials of ${\rm Sing}(X)$ include virtual polynomials in the second layer,
it is enough to show that any linear polynomials appearing in the defining polynomials are virtual polynomials in the second layer. If we assume that a linear polynomial appears in the defining polynomials of ${\rm Sing}(X)$, it will be the irreducible component of a virtual polynomial. By Theorem 2.8, we can see that it is the virtual polynomial of the second layer or weight parameter. This is because, if a linear polynomial that is not a virtual polynomial of the second layer appears as an irreducible component of the virtual polynomial, it must start from the $l$-th layer with one active node and end at the $l+1$-th layer with one active node. This is a weight parameter.
Hence, we reconstruct the input of samples up to scalar multiplication and the weights on the paths from the first layer to the second layer.  \\
Next, We identify the weights on the paths from the $l$-th layer to the $l+1$-th layer by induction on $l$.
By the discussion above, we identified the weights of the first layer, namely $l=1$.
Assume that we identified the weights for the $l$-th layer if $l<k$.
Pick up the polynomials $f$ of the defining inequalities of degree $k$ such that each monomial in $f$ is consist of the weights of $l<k$ except one weight parameter. We can easily see that such weight is on the paths from the $k$-th layer to the $k+1$-th layer by induction on $l$. Hence, we can also obtain the number of the layers. This completes the proof.
\subsection{Proof of Proposition 2.12}
We note that the ambient space of the loss surface is $\mathbb{R}^{N+1}$.
Hence, we can determine the hyperplane going through the points.
This hyperplane will be a linear component of ${\rm Sing}(X)$, and we obtain a sample up to scalar.

\end{document}